\documentclass[11pt,a4paper]{article}
\usepackage[hyperref]{acl2019}
\usepackage{times}
\usepackage{latexsym}

\usepackage{graphicx}

\usepackage{CJK,algorithm,algorithmic,amssymb,amsmath,array,epsfig,graphics,multirow,array,float,subfigure,verbatim,epstopdf}
\usepackage{enumitem}
\usepackage{color,soul}
\usepackage{hhline}
\usepackage{multirow}
\usepackage{xcolor}
\usepackage{hyperref}
\usepackage{tabularx}
\usepackage{bm}


\aclfinalcopy % Uncomment this line for the final submission
\def\aclpaperid{279} %  Enter the acl Paper ID here

\usepackage{xparse}
\definecolor{fl1}{RGB}{246,119,118}
\NewDocumentCommand{\heng}{ mO{} }{\textcolor{red}{\textsuperscript{\textit{Heng}}\textsf{\textbf{\small[#1]}}}}
\NewDocumentCommand{\lifu}{ mO{} }{\textcolor{blue}{\textsuperscript{\textit{Lifu}}\textsf{\textbf{\small[#1]}}}}
\NewDocumentCommand{\zhiying}{ mO{} }{\textcolor{orange}{\textsuperscript{\textit{Zhiying}}\textsf{\textbf{\small[#1]}}}}
\NewDocumentCommand{\qingyun}{ mO{} }{\textcolor{purple}{\textsuperscript{\textit{Qingyun}}\textsf{\textbf{\small[#1]}}}}
\NewDocumentCommand{\mohit}{ mO{} }{\textcolor{cyan}{\textsuperscript{\textit{Mohit}}\textsf{\textbf{\small[#1]}}}}

\title{PaperRobot: Incremental Draft Generation of Scientific Ideas}

\author{
Qingyun Wang$^{1}$, \ Lifu Huang$^{1}$, \ Zhiying Jiang$^1$, \\ \  \textbf{Kevin Knight}$^2$, \ \textbf{Heng Ji}$^{1,3}$, \  \textbf{Mohit Bansal}$^4$, \ \textbf{Yi Luan}$^5$\\
$^{1}$ Rensselaer Polytechnic Institute $^{2}$ DiDi Labs $^{3}$ University of Illinois at Urbana-Champaign\\
 $^{4}$ University of North Carolina at Chapel Hill $^{5}$ University of Washington \\
{\tt kevinknight@didiglobal.com}, {\tt hengji@illinois.edu}\\
}

\date{}

\begin{document}
\maketitle

\begin{abstract}

We present a \emph{PaperRobot} who performs as an automatic research assistant by (1) conducting deep understanding of a large collection of human-written papers in a target domain and constructing comprehensive background knowledge graphs (KGs);
(2) creating new ideas by predicting links from the background KGs, by combining graph attention and contextual text attention; (3) incrementally writing some key elements of a new paper based on memory-attention networks: from the input title along with predicted related entities to generate a paper abstract, from the abstract to generate conclusion and future work, and finally from future work to generate a title for a follow-on paper.
Turing Tests, where a biomedical domain expert is asked to compare a system output and a human-authored string, show \emph{PaperRobot} generated abstracts, conclusion and future work sections, and new titles are chosen over human-written ones up to 30\%, 24\% and 12\% of the time, respectively.\footnote{The programs, data and resources are publicly available for research purpose at: \url{https://github.com/EagleW/PaperRobot}}

\end{abstract}

\section{Introduction}

Our ambitious goal is to speed up scientific discovery and production by building a \emph{PaperRobot},
who addresses three main tasks as follows.

\textbf{Read Existing Papers.} Scientists now find it difficult to keep up with the overwhelming amount of papers. For example, in the biomedical domain, on average more than 500K papers are published every year\footnote{\url{http://dan.corlan.net/medline-trend/language/absolute.html}}, and more than 1.2 million new papers are published in 2016 alone, bringing the total number of papers to over 26 million~\cite{van2014scientists}. However, human's reading ability keeps almost the same across years. In 2012, US scientists estimated that they read, on average, only 264 papers per year (1 out of 5000 available papers), which is, statistically, not different from what they reported in an identical survey last conducted in 2005. \emph{PaperRobot} automatically reads existing papers to build background knowledge graphs (KGs), in which nodes are entities/concepts and edges are the relations between these entities
(Section~\ref{subsec:extraction}).

\begin{figure}[!hbt]
\centering
\includegraphics[width=\linewidth]{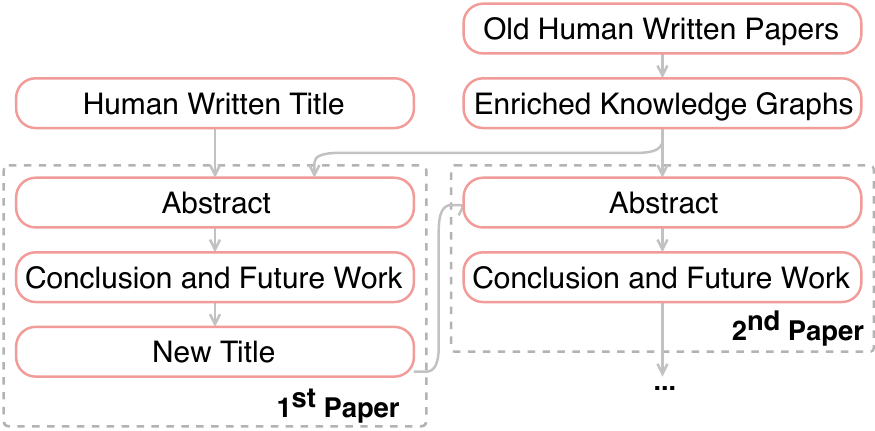}
\caption{PaperRobot Incremental Writing}
\label{img:writing}
\end{figure}

\begin{figure*}[!hbt]
\centering
\includegraphics[width=\linewidth]{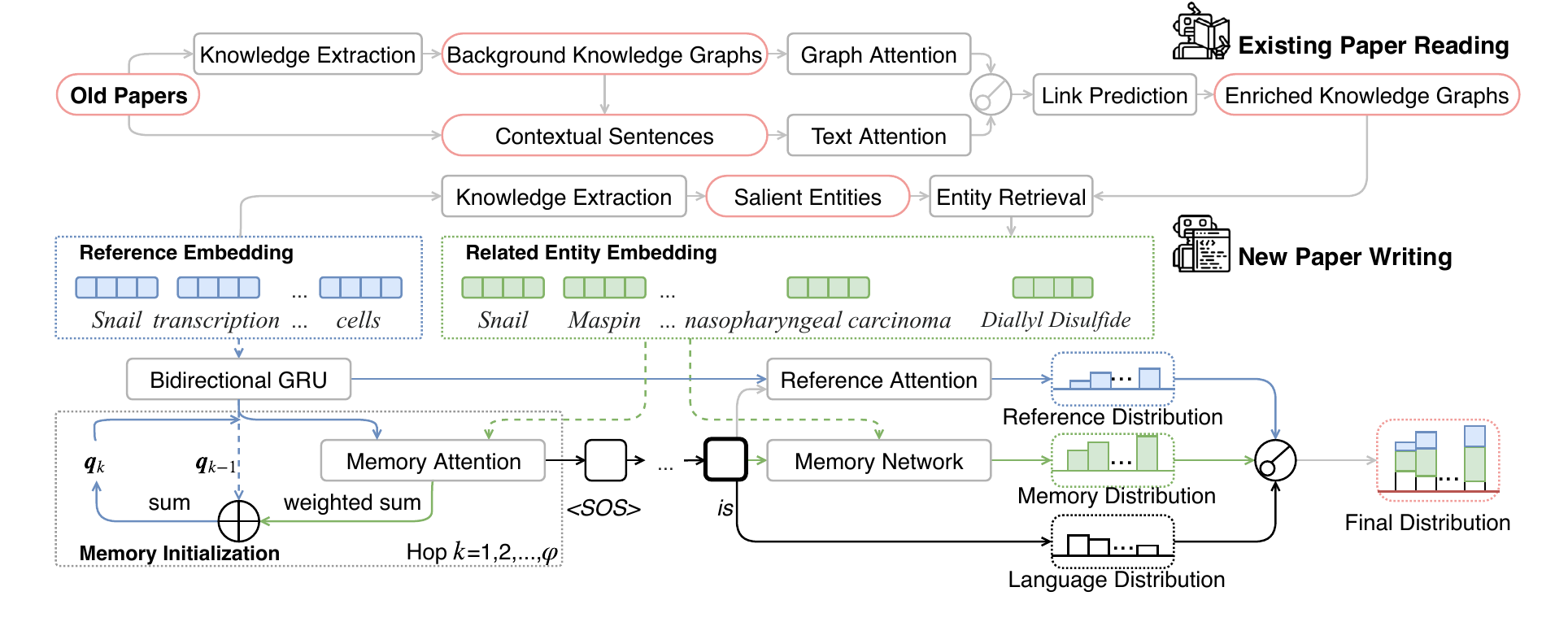}
\caption{PaperRobot Architecture Overview}
\label{img:overview}
\end{figure*}

\textbf{Create New Ideas.} Scientific discovery can be considered as creating new nodes or links in the knowledge graphs. Creating new nodes usually means discovering new entities (e.g., new proteins) through a series of real laboratory experiments, which is probably too difficult for \emph{PaperRobot}. In contrast, creating new edges is easier to automate using the background knowledge graph as the starting point. ~\newcite{Foster2015} shows that more than 60\% of 6.4~million papers in biomedicine and chemistry are about incremental work. This inspires us to  automate the incremental creation of new ideas and hypotheses by predicting new links in background KGs.
In fact, when there is more data available, we can construct larger and richer background KGs for more reliable link prediction. Recent work~\cite{Ji2015} successfully mines strong relevance between drugs and diseases from biomedical papers based on KGs constructed from weighted co-occurrence.
We propose a new entity representation that combines KG structure and unstructured contextual text for link prediction (Section~\ref{subsec:prediction}).

\textbf{Write a New Paper about New Ideas.} The final step is to communicate the new ideas to the reader clearly,
which is a very difficult thing to do; many scientists are, in fact, bad writers~\cite{Pinker2014}. Using a novel memory-attention network architecture, \emph{PaperRobot} automatically writes a new paper abstract about an input title along with predicted related entities,
then further writes conclusion and future work based on the abstract, and finally predicts a new title for a future follow-on paper, as shown in Figure~\ref{img:writing} (Section~\ref{subsec:generation}).

We choose biomedical science as our target domain due to the sheer volume of available papers.
Turing tests show that \emph{PaperRobot}-generated output strings are sometimes chosen over human-written ones;
and most paper abstracts only require minimal edits from domain experts to become highly informative and coherent.

\begin{figure*}[!hbt]
\centering
\includegraphics[width=\linewidth]{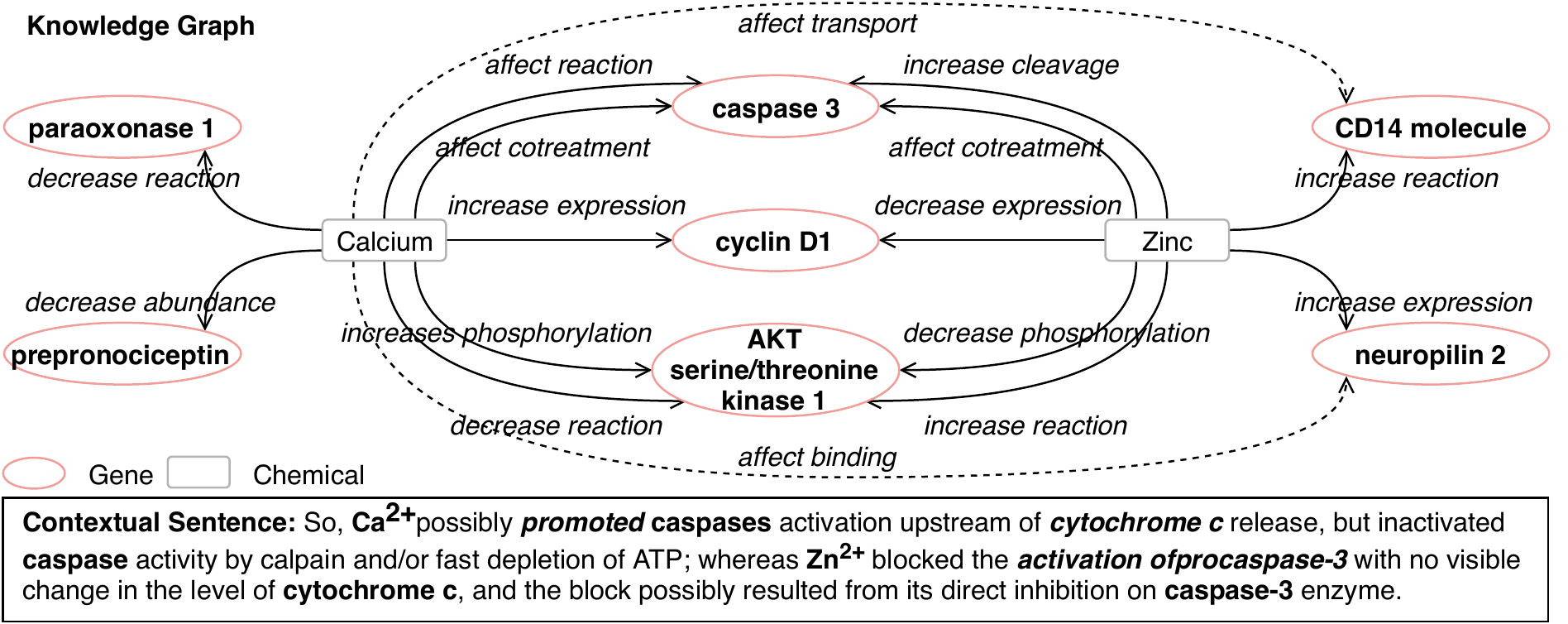}
\caption{Biomedical Knowledge Extraction and Link Prediction Example (dash lines are predicted links)}
\label{img:prediction}
\end{figure*}
\section{Approach}
\subsection{Overview}
The overall framework of \emph{PaperRobot} is illustrated in Figure~\ref{img:overview}. A walk-through example produced from this whole process is shown in Table~\ref{table:walkthrough}.
In the following subsections, we will elaborate on the algorithms for each step.

\subsection{Background Knowledge Extraction}
\label{subsec:extraction}

From a massive collection of existing biomedical papers,
we extract entities and their relations to construct background knowledge graphs (KGs).
We apply an entity mention extraction and linking system~\cite{wei2013pubtator} to extract mentions of three entity types (\textbf{Disease}, \textbf{Chemical} and \textbf{Gene}) which are the core data categories in the Comparative Toxicogenomics Database (CTD)~\cite{davis2016comparative},
and obtain a Medical Subject Headings (MeSH) Unique ID for each mention. Based on the MeSH Unique IDs, we further link all entities to the CTD and extract 133 subtypes of relations such as \textbf{Marker/Mechanism}, \textbf{Therapeutic}, and \textbf{Increase Expression}. Figure~\ref{img:prediction} shows an example.

\subsection{Link Prediction}
\label{subsec:prediction}

After constructing the initial KGs from existing papers, we perform link prediction to enrich them.
Both contextual text information and graph structure are important to represent an entity, thus we combine them to generate a rich representation for each entity.
Based on the entity representations, we determine whether any two entities are semantically similar, and if so, we propagate the neighbors of one entity to the other.
For example, in Figure~\ref{img:prediction}, because \emph{Calcium} and \emph{Zinc} are similar in terms of contextual text information and graph structure, we predict two new neighbors for \emph{Calcium}: \emph{CD14 molecule} and \emph{neuropilin 2} which are neighbors of \emph{Zinc} in the initial KGs.

We formulate the initial KGs as a list of tuples numbered from $0$ to $\kappa$. Each tuple $(e_i^{h}, r_i, e_i^{t}) $ is composed of a head entity $e_i^{h}$, a tail entity $e_i^{t}$,  and their relation $r_i$. Each entity $e_i$ may be involved in multiple tuples and its one-hop connected neighbors are denoted as $N_{e_i}=[n_{i1}, n_{i2}, ...]$. $e_i$ is also associated with a context description $s_i$ which is randomly selected from the sentences where $e_i$ occurs. We randomly initialize vector representations $\boldsymbol{e_i}$ and  $\boldsymbol{r_i}$ for $e_i$ and $r_i$ respectively.

\noindent\textbf{Graph Structure Encoder}
To capture the importance of each neighbor's feature to $e_i$,
we perform self-attention~\cite{velickovic2018graph} and compute a weight distribution over $N_{e_i}$:
\begin{align*}
\boldsymbol{e}_{i}^{'} &= \boldsymbol{W}_{e}\boldsymbol{e}_{i}, \quad
\boldsymbol{n}_{ij}^{'} = \boldsymbol{W}_{e}\boldsymbol{n}_{ij}
\\
c_{ij} &= \mathrm{LeakyReLU}(\boldsymbol{W}_f(\boldsymbol{e}_{i}^{'}\oplus\boldsymbol{n}_{ij}^{'}))
\\
\boldsymbol{c}_{i}^{'} &= \mathrm{Softmax}(\boldsymbol{c}_{i})
\end{align*}
where $\boldsymbol{W}_e$ is a linear transformation matrix applied to each entity.
$\boldsymbol{W}_f$ is the parameter for a single layer feedforward network. $\oplus$ denotes the concatenation operation between two matrices.
Then we use $\boldsymbol{c}_{i}^{'}$ and $N_{e_i}$ to compute a structure based context representation of $
\boldsymbol{\epsilon}_{i} = \sigma\left(\sum c_{ij}^{'}\boldsymbol{n}_{ij}^{'}\right)$, where $n_{ij}\in N_{e_i}$ and $\sigma$ is Sigmoid function.

In order to capture various types of relations between $e_i$ and its neighbors, we further perform multi-head attention on each entity, based on multiple linear transformation matrices.
Finally, we get a structure based context representation $\tilde{\boldsymbol{e}}_{i} = [\boldsymbol{\epsilon}_{i}^{0}\oplus...\oplus\boldsymbol{\epsilon}_{i}^{M}]$, where $\boldsymbol{\epsilon}_{i}^{m}$ refers to the context representation obtained with the $m$-$\mathrm{th}$ head, and $\boldsymbol{\tilde{e}}_{i}$ is the concatenated representation based on the attention of all $M$ heads.

\noindent\textbf{Contextual Text Encoder}
Each entity $e$ is also associated with a context sentence $[w_1, ..., w_l]$. To incorporate the local context information, we first apply a bi-directional long short-term memory (LSTM) \cite{graves2005framewise} network to get the encoder hidden states $\boldsymbol{H}_{s} = [\boldsymbol{h}_{1}, ..., \boldsymbol{h}_{l}]$, where $\boldsymbol{h}_{i}$ represents the hidden state of $w_i$.
Then we compute a bilinear attention weight for each word $w_i$: $\mu_i = \boldsymbol{e}^{\top}\boldsymbol{W}_s \boldsymbol{h}_{i},
\bm{\mu}^{'} = \mathrm{Softmax}(\bm{\mu})$, where $\boldsymbol{W}_s$ is a bilinear term. We finally get the context representation  $\hat{\boldsymbol{e}} = \bm{\mu}^{'\top}{\boldsymbol{h}_i}$. 

\begin{table*}[ht!]
\small
\begin{tabularx}{\linewidth}{|>{\hsize=0.4\hsize}X|>{\arraybackslash\hsize=1.4\hsize}X|>{\arraybackslash\hsize=1.18\hsize}X|>{\arraybackslash\hsize=1.02\hsize}X|}\hline
Title         & \multicolumn{3}{>{\hsize=\dimexpr 3.62\hsize+2\tabcolsep+\arrayrulewidth}X|}{\textbf{Snail} transcription factor negatively regulates \textbf{maspin tumor suppressor} in \textbf{human prostate cancer} cells}          \\ \hline
Entities         & \multicolumn{3}{>{\hsize=\dimexpr 2.7\hsize+2\tabcolsep+\arrayrulewidth}X|}{
Related: \textbf{nasopharyngeal carcinoma}; diallyl disulfide
}          \\ \hline
Output & \multicolumn{1}{c|}{\textbf{Human~\cite{neal2012snail}}} & \multicolumn{1}{c|}{\textbf{System}} & \multicolumn{1}{c|}{\textbf{Post-edited by Human}}\\ \hline

Abstract      & Background: \textbf{Maspin}, a putative \textbf{tumor suppressor} that is down-regulated in breast and \textbf{prostate cancer}, has been associated with decreased cell motility. \textbf{Snail} transcription factor is a zinc finger protein that is increased in breast cancer and is associated with increased tumor motility and invasion by induction of epithelial-mesenchymal transition (EMT). We investigated the molecular mechanisms by which \textbf{Snail} increases \textbf{tumor} motility and invasion utilizing \textbf{prostate cancer} cells. Methods: Expression levels were analyzed by \textbf{RT-PCR} and \textbf{western blot} analyses. Cell motility and invasion assays were performed , while \textbf{Snail} regulation and binding to \textbf{maspin} promoter was analyzed by luciferase reporter and chromatin immunoprecipitation (ChIP) assays. Results: \textbf{Snail} protein expression was higher in different \textbf{prostate cancer} cells lines as compared to normal prostate epithelial cells.
& Background: \textbf{Snail} is a multifunctional protein that plays an important role in the pathogenesis of \textbf{prostate cancer}. \textit{However}, it has been shown \textit{to be} associated with poor prognosis. The purpose of this study \textit{was} to investigate the effect of \textit{negatively} on the expression of \textbf{\textit{maspin}} in\textbf{ human nasopharyngeal carcinoma} cell lines. Methods: \textit{Quantitative real-time PCR} and western blot analysis were used to determine \textit{whether the demethylating agent was investigated by quantitative \textbf{RT-PCR} (qRT-PCR) and\textbf{ Western blotting}.} Results showed that the binding protein plays a significant role in the regulation of \textbf{tumor} growth and progression.
& Background: \textbf{Snail} is a multifunctional protein that plays an important role in the pathogenesis of \textbf{prostate cancer}. It has been shown associated with poor prognosis. The purpose of this study is to investigate the negative effect of on the expression of \textbf{Maspin} in\textbf{ human nasopharyngeal carcinoma} cell lines. Methods: Quantitative \textbf{RT-PCR} \textit{(qRT-PCR)} and \textbf{western blot analyses} were used to determine \textit{correlation of the two proteins’ expressions.} Results showed that the binding protein plays a significant role in the regulation of \textbf{tumor} growth and progression.
\\ \hline
\centering  Conclusion and Future work   &   Collectively, our results indicate for the first time that \textbf{Snail} can negatively regulate \textbf{maspin} through direct promoter repression resulting in increased migration and invasion in prostate \textbf{cancer} cells. This study reveals a novel mechanism of how \textbf{Snail} may function and show the importance of therapeutic targeting of \textbf{Snail} signaling in future.
& In summary, our study demonstrates that \textbf{Snail} negatively \textit{inhibited} the expression of \textbf{Maspin} in \textbf{human nasopharyngeal carcinoma} cell lines \textit{and in vitro}. Our results indicate that \textit{the combination of the demethylating agent} might be a potential therapeutic target for the treatment of \textbf{prostate cancer}.
& In summary, our study \textit{in vitro} demonstrates that \textbf{Snail} negatively \textit{inhibits} the expression of \textbf{Maspin} in \textbf{human nasopharyngeal carcinoma} cell lines. Our results \textit{further} indicate that \textbf{\textit{Maspin}} might be a potential therapeutic target for the treatment of \textbf{prostate cancer}.
\\ \hline
New Title &    Role of \textbf{maspin} in cancer \cite{berardi2013role}
& The role of \textbf{\textit{nasopharyngeal carcinoma}} in the rat model of \textbf{\textit{prostate cancer}} cells
& The role of \textit{\textbf{Maspin}} in the rat model of \textbf{\textit{nasopharyngeal carcinoma}} cells
\\ \hline
\end{tabularx}
\caption{\label{table:walkthrough}Comparison of Human and System Written Paper Elements (bold words are topically related entities; italic words show human edits)
}

\end{table*}
\noindent\textbf{Gated Combination}
To combine the graph-based representation $\tilde{\boldsymbol{e}}$ and local context based representations $\hat{\boldsymbol{e}}$, we design a gate function to balance these two types of information:
\begin{align*}
\boldsymbol{g}_{e} =\sigma (\tilde{\boldsymbol{g}}_e), \quad
\boldsymbol{e} = \boldsymbol{g}_{e}\odot\tilde{\boldsymbol{e}} + (1-\boldsymbol{g}_{e})\odot\hat{\boldsymbol{e}}
\end{align*}
where $\boldsymbol{g}_{e}$ is an entity-dependent gate function of which each element is in $[0, 1]$, $\tilde{\boldsymbol{g}}_e$ is a learnable parameter for each entity $e$, $\sigma$ is a Sigmoid function, and $\odot$ is an element-wise multiplication.

\noindent\textbf{Training and Prediction}
To optimize both entity and relation representations, following TransE~\cite{bordes2013translating}, we assume the relation between two entities can be interpreted as translations operated on the entity representations, namely $\boldsymbol{h}+\boldsymbol{r}\approx\boldsymbol{t}$ if $(h, r, t)$ holds. Therefore, for each tuple $(e_{i}^{h},r_{i},e_{i}^{t})$, we can compute their distance score: $F(e_{i}^{h},r_{i},e_{i}^{t}) = \parallel \boldsymbol{e}_{i}^{h} + \boldsymbol{r}_{i} - \boldsymbol{e}_{i}^{t}\parallel_{2}^2$.
We use marginal loss to train the model:
\begin{align*}
\begin{split}
Loss = \sum_{(e_{i}^{h},r_{i},e_{i}^{t})\in K}\sum_{(\bar{e}_{i}^{h}, \bar{r}_{i}, \bar{e}_{i}^{t})\in\bar{K}}\mathrm{max}(0, \\
\gamma + F(e_{i}^{h},r_{i},e_{i}^{t}) -  F(\bar{e}_{i}^{h},\bar{r}_{i},\bar{e}_{i}^{t}))
\end{split}
\end{align*}
where $(e^{h},r,t^h)$ is a positive tuple and $(\bar{e}^h,\bar{r}^h,\bar{t}^h)$ is a negative tuple, and $\gamma$ is a margin. The negative tuples are generated by either replacing the head or the tail entity of positive tuples with a randomly chosen different entity.

After training, for each pair of indirectly connected entities $e_i$, $e_j$ and a relation type $r$, we compute a score $y$ to indicate the probability that $(e_i, r, e_j)$ holds, and obtain an enriched knowledge graph $\widetilde{K}=[(e_{\kappa+1}^{h}, r_{\kappa+1}, e_{\kappa+1}^{t}, y_{\kappa+1})...]$.

\subsection{New Paper Writing}
\label{subsec:generation}

In this section, we use title-to-abstract generation as a case study to describe the details of our paper writing approach. Other tasks (abstract-to-conclusion and future work, and conclusion and future work-to-title) follow the same architecture.

Given a reference title $\tau=[w_1, ..., w_l]$,
we apply the knowledge extractor (Section~\ref{subsec:extraction}) to extract entities from $\tau$. For each entity, we retrieve a set of related entities from the enriched knowledge graph $\widetilde{K}$ after link prediction. We rank all the related entities by confidence scores and select up to 10 most related entities $E_{\tau} = [e_{1}^{\tau}, ..., e_{v}^{\tau}]$. Then we feed $\tau$ and $E_{\tau}$ together into the paper generation framework as shown in Figure~\ref{img:overview}. The framework is based on a hybrid approach of a Mem2seq model~\citep{P18-1136} and a pointer generator~\cite{copy16,hybridp17}. It allows us to balance three types of sources for each time step during decoding: the probability of generating a token from the entire word vocabulary based on language model, the probability of copying a word from the reference title, such as \textit{regulates} in Table~\ref{table:walkthrough}, and the probability of incorporating a related entity, such as \textit{Snail} in Table~\ref{table:walkthrough}. The output is a paragraph $Y=[y_1,...,y_o].$\footnote{During training, we truncate both of the input and the output to around 120 tokens to expedite training. We label the words with frequency $<5$ as Out-of-vocabulary.}

\noindent\textbf{Reference Encoder}
For each word in the reference title, we randomly embed it into a vector and obtain $\bm{\tau}=[\boldsymbol{w}_1,..., \boldsymbol{w}_l]$. Then, we apply a bi-directional Gated Recurrent Unit (GRU) encoder~\cite{cho2014learning} on $\bm{\tau}$ to produce the encoder hidden states $\boldsymbol{H}=[\boldsymbol{h}_1, ..., \boldsymbol{h}_l]$.

\noindent\textbf{Decoder Hidden State Initialization}
Not all predicted entities are equally relevant to the title. For example, for the title in Table~\ref{img:overview}, we predict multiple related entities including \textit{nasopharyngeal carcinoma} and \textit{diallyl disulfide}, but \textit{nasopharyngeal carcinoma} is more related because \textit{nasopharyngeal carcinoma} is also a cancer related to \textit{snail transcription factor}, while \textit{diallyl disulfide} is less related because \textit{diallyl disulfide}'s anticancer mechanism is not closely related to \textit{maspin tumor suppressor}. We propose to apply memory-attention networks to further filter the irrelevant ones.
Recent approaches~\cite{sukhbaatar2015end,P18-1136}
show that compared with soft-attention, memory-based multihop attention is able to refine the attention weight of each memory cell to the query multiple times, drawing better correlations. Therefore, we apply a multihop attention mechanism to generate the initial decoder hidden state.

Given the set of related entities $E = [e_{1}, ..., e_{v}]$, we randomly initialize their vector representation $\boldsymbol{E} = [\boldsymbol{e}_{1}, ..., \boldsymbol{e}_{v}]$ and store them in memories. Then we use the last hidden state of reference encoder $\boldsymbol{h}_l$ as the first query vector $\boldsymbol{q}_0$, and iteratively compute the attention distribution over all memories and update the query vector:
\begin{align*}
p_{ki}&= \boldsymbol{\nu}^{\top}_{k}\tanh{\left(\boldsymbol{W}_{q}^{k}\boldsymbol{q}_{k-1} + \boldsymbol{U}_{e}^{k}\boldsymbol{e}_i + \boldsymbol{b}_{k}\right)}\\
\boldsymbol{q}_k &=\boldsymbol{p}_{k}^{\top}\boldsymbol{e} +\boldsymbol{q}_{k-1}
\end{align*}
where $k$ denotes the $k$-th hop among $\varphi$ hops in total.\footnote{We set $\varphi=3$ since it performs the best on the development set.}
After $\varphi$ hops, we obtain $\boldsymbol{q}_{\varphi}$ and take it as the initial hidden state of the GRU decoder.

\noindent\textbf{Memory Network}
To better capture the contribution of each entity $e_j$ to each decoding output, at each decoding step $i$, we compute an attention weight for each entity and apply a memory network to refine the weights multiple times. We take the hidden state $\tilde{\boldsymbol{h}}_{i}$ as the initial query $\tilde{\boldsymbol{q}}_0 = \tilde{\boldsymbol{h}}_{i}$ and iteratively update it:
\begin{align*}
\tilde{p}_{kj}&= \nu^{\top}_{k}\tanh{\left(\widetilde{\boldsymbol{W}}_{\tilde{q}}^{k}\tilde{\boldsymbol{q}}_{k-1} + \widetilde{\boldsymbol{U}}_{e}^{k}\boldsymbol{e}_j + \boldsymbol{W}_{\hat{c}}\hat{c}_{ij} + b_k\right)}\\
\boldsymbol{u}_{ik}&=\boldsymbol{\tilde{p}}_{k}^{'\top} \boldsymbol{e}_j, \quad
\tilde{\boldsymbol{q}}_k = \boldsymbol{u}_{ik} +\tilde{\boldsymbol{q}}_{k-1}
\end{align*}
where $\hat{\boldsymbol{c}}_{ij}=\sum_{m=0}^{i-1}\boldsymbol{\beta}_{mj}$ is an entity coverage vector and $\boldsymbol{\beta}_{i}$ is the attention distribution of last hop $\boldsymbol{\beta}_{i}=\boldsymbol{\tilde{p}}_{\psi}^{'}$, and $\psi$ is the total number of hops. We then obtain a final memory based context vector for the set of related entities $\boldsymbol{\chi}_{i}=\boldsymbol{u}_{i\psi}$.

\noindent\textbf{Reference Attention} Our reference attention is similar to~\cite{atten15,hybridp17}, which aims to capture the contribution of each word in the reference title to the decoding output. At each time step $i$, the decoder receives the previous word embedding and generate decoder state $\tilde{\boldsymbol{h}}_i$, the attention weight of each reference token is computed as:
\begin{align*}
&\alpha_{ij} = \boldsymbol{\varsigma}^{\top}\tanh{\left(\boldsymbol{W}_h \tilde{\boldsymbol{h}}_i + \boldsymbol{W}_{\tau} \boldsymbol{h}_j +  \boldsymbol{W}_{\tilde{c}}\boldsymbol{\tilde{c}}_{ij} + \boldsymbol{b}_{\tau}\right)}\\
&\boldsymbol{\alpha}_{i}^{'} =\mathrm{Softmax}\left(\boldsymbol{\alpha}_i\right); \quad
\boldsymbol{\phi}_{i}=\boldsymbol{\alpha}_{i}^{'\top}\boldsymbol{h}_j
\end{align*} $\boldsymbol{\tilde{c}}_{ij}=\sum_{m=0}^{i-1}\alpha _{mj}$ is a reference coverage vector, which is the sum of attention distributions over all previous decoder time steps to reduce repetition~\cite{hybridp17}. $\boldsymbol{\phi}_{i}$ is the reference context vector.

\begin{table*}[!htb]
\centering
\small
\begin{tabularx}{\linewidth}{|>{\hsize=0.7\hsize}X|>{\centering\arraybackslash\hsize=0.6\hsize}X|>{\centering\arraybackslash\hsize=1.5\hsize}X|>{\centering\arraybackslash\hsize=1.3\hsize}X|>{\centering\arraybackslash\hsize=0.8\hsize}X|>{\centering\arraybackslash\hsize=1.1\hsize}X|}
\hline
\multirow{2}{*}{\textbf{Dataset}}&\multicolumn{3}{c|}{\textbf{\# papers}}& \multirow{2}{*}{\parbox{\dimexpr 1\hsize}{\centering\textbf{\# avg entities in Title / paper}}}&\multirow{2}{*}{\parbox{\dimexpr 1\hsize}{\centering\textbf{\# avg predicted related entities / paper}}}       \\ \cline{2-4}
&\textbf{Title-to-Abstract }&\textbf{Abstract-to-Conclusion and Future work}&\textbf{Conclusion and Future work-to-Title}&&\\\hline
Training&22,811&22,811&15,902&4.8 & -\\ \hline
Development&2,095&2,095&2,095& 5.6& 6.1\\ \hline
Test&2,095&2,095&2,095&5.7 & 8.5\\ \hline
\end{tabularx}
\caption{Paper Writing Statistics\label{table:write}}
\end{table*}
\begin{table*}[!htb]
\centering
\small
\begin{tabularx}{\linewidth}{|>{\hsize=2.6\hsize}X|>{\centering\arraybackslash\hsize=0.7\hsize}X|>{\centering\arraybackslash\hsize=0.7\hsize}X|>{\centering\arraybackslash\hsize=0.8\hsize}X|>{\centering\arraybackslash\hsize=0.8\hsize}X|>{\centering\arraybackslash\hsize=0.7\hsize}X|>{\centering\arraybackslash\hsize=0.7\hsize}X|}
\hline
\multirow{2}{*}{\textbf{Model}} & \multicolumn{2}{c|}{\textbf{Title-to-Abstract}} & \multicolumn{2}{>{\centering\arraybackslash\hsize=1,8\hsize}X|}{\textbf{Abstract-to-Conclusion and Future Work}} & \multicolumn{2}{>{\centering\arraybackslash\hsize=1.6\hsize}X|}{\textbf{Conclusion and Future Work-to-Title}} \\ \cline{2-7}
 & \textbf{Perplexity} & \textbf{METEOR} & \textbf{Perplexity}  & \textbf{METEOR} & \textbf{Perplexity} & \textbf{METEOR}  \\ \hline
Seq2seq~\cite{atten15} & 19.6 & 9.1 & 44.4  & 8.6 & 49.7 & 6.0  \\ \hline
Editing Network~\citep{P18-2042}& 18.8 &  9.2 & 30.5 & 8.7 & 55.7 &5.5 \\ \hline
Pointer Network~\cite{hybridp17} &  146.7  &8.5    & 74.0 & 8.1 & 47.1  &6.6 \\ \hline
Our Approach (-Repetition Removal)& 13.4 &12.4 & 24.9 & \textbf{12.3} & 31.8 & 7.4\\ \hline
Our Approach & \textbf{11.5} &\textbf{13.0} & \textbf{18.3} & 11.2 & \textbf{14.8} & \textbf{8.9}\\ \hline
\end{tabularx}
\caption{Automatic Evaluation on Paper Writing for Diagnostic Tasks (\%). The Pointer Network can be viewed as removing memory network part from our approach without repetition removal.  \label{table:methodcomparison}}
\end{table*}

\noindent\textbf{Generator}
% \qingyun{changed}
For a particular word $w$, it may occur multiple times in the reference title or in multiple related entities. Therefore, at each decoding step $i$, for each word $w$, we aggregate its attention weights from the reference attention and memory attention distributions:
$P_\tau^{i} = \sum_{m|w_m=w} \boldsymbol{\alpha}_{im}^{'}$ and $P_e^{i} = \sum_{m|w\in e_m} \boldsymbol{\beta}_{im}$ respectively.
In addition, at each decoding step $i$, each word in the vocabulary may also be generated with a probability according to the language model. The probability is computed from the decoder state $\tilde{\boldsymbol{h}}_i$, the reference context vector $\boldsymbol{\phi}_{i}$, and the memory context vector $\boldsymbol{\chi}_{i}$: $
P_{gen} = \mathrm{Softmax}(\boldsymbol{W}_{gen}[\tilde{\boldsymbol{h}}_i;\boldsymbol{\phi}_{i};\boldsymbol{\chi}_{i}]+\boldsymbol{b}_{gen})$, where $\boldsymbol{W}_{gen}$ and $b_{gen}$ are learnable parameters.
To combine $P_\tau$, $P_e$ and $P_{gen}$, we compute a gate $\boldsymbol{g}_{\tau}$ as a soft switch between generating a word from the vocabulary and copying words from the reference title $\tau$ or the related entities $E$: $\boldsymbol{g}_p = \sigma( \boldsymbol{W}^{\top}_p \tilde{\boldsymbol{h}}_i + \boldsymbol{W}^{\top}_z \boldsymbol{z}_{i-1} + \boldsymbol{b}_p)
$, where $\boldsymbol{z}_{i-1}$ is the embedding of the previous generated token at step $i-1$. $\boldsymbol{W}_p$, $\boldsymbol{W}_z$, and $\boldsymbol{b}_p$ are learnable parameters, and $\sigma$ is a Sigmoid function.
We also compute a gate $\tilde{\boldsymbol{g}}_p$ as a soft switch between copying words from reference text and the related entities:
$
\tilde{\boldsymbol{g}}_p = \sigma( \boldsymbol{W}^{\top}_\phi \boldsymbol{\phi}_{i} +\boldsymbol{W}^{\top}_\chi \boldsymbol{\chi}_{i} +  \tilde{\boldsymbol{b}}_p)
$,
where $\boldsymbol{W}_{\phi}$, $\boldsymbol{W}_{\chi}$, and $\tilde{\boldsymbol{b}}_p$ are learnable parameters.

The final probability of generating a token $z$ at decoding step $i$ can be computed by:
\begin{align*}
P(z_i) &= \boldsymbol{g}_p P_{gen}+(1-\boldsymbol{g}_p)\left(\tilde{\boldsymbol{g}}_p P_\tau +(1-\tilde{\boldsymbol{g}}_p)P_e\right)
\end{align*}

The loss function, combined with the coverage loss~\citep{hybridp17} for both reference attention and memory distribution, is presented as:
\begin{align*}
\begin{split}
Loss &= \sum\nolimits_{i} -\log{P(z_i)}+\lambda\sum\nolimits_{i}\left(\min{\left(\alpha_{ij},\boldsymbol{\tilde{c}}_{ij}\right)}\right.
\\
&\mathrel{\phantom{=}}\left.+\min{\left(\beta_{ij},\hat{\boldsymbol{c}}_{ij}\right)}\right)
\end{split}
\end{align*}
where $P(z_i)$ is the prediction probability of the ground truth token $z_i$, and $\lambda$ is a hyperparameter.

\noindent\textbf{Repetition Removal}
Similar to many other long text generation tasks~\cite{E17-2047}, repetition remains a major challenge~\cite{Foster2007,xie2017neural}.
In fact, 11\% sentences in human written abstracts include repeated entities, which may mislead the language model.
Following the coverage mechanism proposed by~\cite{P16-1008,hybridp17}, we use a coverage loss to avoid any entity in reference input text or related entity receiving attention multiple times.
We further design a new and simple masking method to remove repetition during the test time. We apply beam search with beam size 4 to generate each output, if a word is not a stop word or punctuation and it is already generated in the previous context, we will not choose it again in the same output.

\section{Experiment}

\subsection{Data}
We collect biomedical papers from the PMC Open Access Subset.\footnote{\url{ftp://ftp.ncbi.nlm.nih.gov/pub/pmc/oa_package/}} To construct ground truth for new title prediction, if a human written paper \textit{A} cites a paper \textit{B}, we assume the title of \textit{A} is generated from \textit{B}'s conclusion and future work session.
We construct background knowledge graphs from 1,687,060 papers which include 30,483 entities and 875,698 relations.
Tables~\ref{table:write} shows the detailed data statistics. The hyperparameters of our model are presented in the Appendix.

\begin{table*}[!htbp]
\centering
\small
\begin{tabularx}{\linewidth}{|>{\hsize=0.8\hsize}X|>{\centering\arraybackslash\hsize=1.6\hsize}X|>{\centering\arraybackslash\hsize=0.8\hsize}X|>{\centering\arraybackslash\hsize=1\hsize}X|>{\centering\arraybackslash\hsize=1\hsize}X|>{\centering\arraybackslash\hsize=0.8\hsize}X|}
\hline
\textbf{Task}&\multicolumn{2}{c|}{\textbf{Input} } &\textbf{Output}                                      & \textbf{Domain Expert} & \textbf{Non-expert}  \\ \hline
%%%%%%%%%%%%
\multirow{7}{*}{End-to-End}&\multirow{2}{*}{Human Title} &   Different									     & \multirow{2}{*}{\centering Abstract (1st)}         &  10&        \textbf{30}                                   \\ \cline{3-3}\cline{5-6}
&                                         &    Same&                                  &  \textbf{30}&         16                           \\ \cline{2-6}
%%%%%%%%%%%%
&\multirow{2}{*}{\parbox{\dimexpr 1.1\hsize}{\centering System Abstract}}   &Different& \multirow{2}{*}{\parbox{\dimexpr 1\hsize}{\centering Conclusion and Future work}} 		         &   \textbf{12}&        		0            \\\cline{3-3}\cline{5-6}
&                        &  Same&                                	       									         & 8&         		8                                    \\ \cline{2-6}
%%%%%%%%%%%%
& \multirow{2}{*}{\parbox{\dimexpr 1\hsize}{\centering System Conclusion and Future work}}     &  Different  	 & \multirow{2}{*}{\parbox{\dimexpr 1\hsize}{\centering Title}}         &   \textbf{12}&           2                           \\ \cline{3-3}\cline{5-6}
&                                         						& Same&   	 		                              &12&        \textbf{25}                                 \\ \cline{2-6}
%%%%%%%%%%%%
&System Title       &Different& \centering Abstract (2nd)   									              &   \textbf{14}&         4                                 \\\hline
%%%%%%%%%%%%
\multirow{4}{*}{Diagnostic}& \multirow{2}{*}{\centering Human Abstract}& 	Different   & \multirow{2}{*}{\parbox{\dimexpr 1\hsize}{\centering Conclusion and Future work}} 	         & 12 &        	\textbf{14}        \\ \cline{3-3}\cline{5-6}
&                              & Same &                           	       									        & \textbf{24}&       20                              \\ \cline{2-6}
%%%%%%%%%%%%
& \multirow{2}{*}{\parbox{\dimexpr 1\hsize}{\centering Human Conclusion and Future work}}     &  Different  	 &\multirow{2}{*}{\centering Title}         & 8  &         \textbf{12}                            \\ \cline{3-3}\cline{5-6}
&                                         						& Same&  		                       	         &2&        \textbf{10}                                    \\ \hline
\end{tabularx}
\caption{Turing Test Human Subject Passing Rates (\%).  Percentages show how often a human judge chooses our system's output over human's when it is mixed with a human-authored string.
If the output strings (e.g., abstracts) are based on the same input string (e.g., title), the Input condition is marked ``Same'', otherwise ``Different''.}
\label{table:turingresult}
\end{table*}
\subsection{Automatic Evaluation}

Previous work~\cite{D16-1230,P16-1094,W15-4640} has proven it to be a major challenge to automatically evaluate long text generation.
Following the story generation work~\cite{P18-1082}, we use METEOR~\cite{denkowski2014meteor} to measure the topic relevance towards given titles and use perplexity to further evaluate the quality of the language model.
The perplexity scores of our model are based on the language model\footnote{\url{https://github.com/pytorch/examples/tree/master/word_language_model}} learned on other PubMed papers (500,000 titles, 50,000 abstracts, 50,000 conclusions and future work) which are not used for training or testing in our experiment.\footnote{The perplexity scores of the language model are in the Appendix.} The results are shown in Table~\ref{table:methodcomparison}. We can see that our framework outperforms all previous approaches.

\begin{table}[!htb]
\small
\begin{tabularx}{\linewidth}{|>{\hsize=1\hsize}X|>{\centering\arraybackslash\hsize=1\hsize}X|>{\centering\arraybackslash\hsize=1.1\hsize}X|>{\centering\arraybackslash\hsize=1.1\hsize}X|>{\centering\arraybackslash\hsize=1.2\hsize}X|>{\centering\arraybackslash\hsize=0.6\hsize}X|}\hline
\textbf{BLEU1} &\textbf{BLEU2} &\textbf{BLEU3} &\textbf{BLEU4} & \textbf{ROUGE} & \textbf{TER} \\\hline
 59.6    &  58.1    & 56.7    & 55.4    &73.3       & 35.2  \\\hline
\end{tabularx}
\caption{Evaluation on Human Post-Editing(\%) \label{table:editingresult}}
\end{table}
\subsection{Turing Test}

Similar to \citep{P18-2042}, we conduct Turing tests by a biomedical expert (non-native speaker) and a non-expert (native speaker). Each human judge is asked to compare a system output and a human-authored string, and select the better one.

Table~\ref{table:turingresult} shows the results on 50 pairs in each setting. We can see that \emph{PaperRobot} generated abstracts are chosen over human-written ones by the expert up to 30\% times, conclusion and future work up to 24\% times, and new titles up to 12\% times. We don't observe the domain expert performs significantly better than the non-expert, because they tend to focus on different aspects - the expert focuses on content (entities, topics, etc.) while the non-expert focuses on the language.

\subsection{Human Post-Editing}
In order to measure the effectiveness of \emph{PaperRobot} acting as a wring assistant, we randomly select 50 paper abstracts generated by the system during the first iteration and ask the domain expert to edit them until he thinks they are informative and coherent.
The BLEU~\cite{Bleu02}, ROUGE~\cite{lin2004rouge} and TER~\cite{snover2006study} scores by comparing the abstracts before and after human editing are presented in Table~\ref{table:editingresult}. It took about 40 minutes for the expert to finish editing 50 abstracts. Table~\ref{table:walkthrough} includes the post-edited example.
We can see that most edits are stylist changes.

\begin{table*}[htb!]
\small
\begin{tabularx}{\linewidth}{|>{\hsize=0.36\hsize}X|>{\arraybackslash\hsize=1.34\hsize}X|>{\arraybackslash\hsize=1.13\hsize}X|>{\arraybackslash\hsize=1.17\hsize}X|}\hline
Output & \multicolumn{1}{c|}{\textbf{Without Memory Networks}} & \multicolumn{1}{c|}{\textbf{Without Link Prediction}} & \multicolumn{1}{c|}{\textbf{Without Repetition Removal}}\\ \hline

Abstract
& Background: \textbf{Snail} has been reported to exhibit a variety of biological functions. In this study, we investigated the effect of negatively on \textbf{maspin} demethylation in \textbf{human prostate cancer} cells. Methods: Quantitative real-time PCR and western blot analysis were used to investigate the effects of the demethylating agent on the expression of the protein kinase (TF) gene promoter. Results: The results showed that the presence of a single dose of 50 $\mu M$ in a dose-dependent manner, whereas the level of the BMP imipramine was significantly higher than that of the control group.
& Background: \textbf{Snail} has been shown to be associated with poor prognosis. In this study, we investigated the effect of negatively on the expression of \textbf{maspin} in \textbf{human prostate cancer} cells. Methods: Cells were treated with a single dose of radiotherapy for 24 h, and was used to investigate the significance of a quantitative factor for the treatment of the disease. Results: The remaining controls showed a significant increase in the G2/M phase of the \textbf{tumor suppressor protein} (p$<$0.05).
& Background: \textbf{Snail} is a major health problem in human malignancies. However, the role of \textbf{Snail} on the expression of \textbf{maspin} in \textbf{human prostate cancer} cells is not well understood. The aim of this study was to investigate the effect of \textbf{Snail} on the expression of \textbf{maspin} in \textbf{human prostate cancer} cells. Methods: The expression of the expression of \textbf{Snail} and \textbf{maspin} was investigated using quantitative RT-PCR and western blot analysis. Results: The remaining overall survival (OS) and overall survival (OS) were analyzed.
\\ \hline
\centering  Conclusion and Future work
& In summary, our study demonstrated that negatively inhibited the expression of the BMP imipramine in \textbf{human prostate cancer} cells. Our findings suggest that the inhibition of \textbf{maspin} may be a promising therapeutic strategy for the treatment.
&  In summary, our results demonstrate that negatively inhibited the expression of \textbf{maspin} in \textbf{human prostate cancer }cells. Our findings suggest that the combination of radiotherapy may be a potential therapeutic target for the treatment of disease.
& In summary, our results demonstrate that \textbf{snail} inhibited the expression of \textbf{maspin} in \textbf{human prostatic} cells. The expression of \textbf{snail} in PC-3 cells by \textbf{snail}, and the expression of \textbf{maspin} was observed in the presence of the expression of \textbf{maspin}.
\\ \hline
New Title
& Protective effects of homolog on \textbf{human breast cancer} cells by inhibiting the Endoplasmic Reticulum Stress
& The role of \textbf{prostate cancer} in \textbf{human breast cancer} cells
& The role of \textbf{maspin} and \textbf{maspin} in human \textbf{breast cancer} cells
\\ \hline
\end{tabularx}
\caption{\label{table:analysis_walk} Ablation Test Results on the Same Title in Table~\ref{table:walkthrough}}
\end{table*}

\begin{table*}[!htb]
\small
\begin{tabularx}{\linewidth}{|>{\hsize=1.1\hsize}X|>{\arraybackslash\hsize=0.9\hsize}X|}\hline
\textbf{Titles}                     & \textbf{Predicted Related Entities}   \\\hline
\textbf{Pseudoachondroplasia/COMP} — translating from the bench to the bedside&osteoarthritis; skeletal dysplasia; thrombospondin-5\\\hline
Role of \textbf{ceramide} in \textbf{diabetes mellitus}: evidence and mechanisms&diabetes insulin ceramide; metabolic disease\\\hline
Exuberant clinical picture of \textbf{Buschke-Fischer-Brauer palmoplantar keratoderma} in bedridden patient&neoplasms; retinoids; autosomal dominant disease\\\hline
Relationship between \textbf{serum adipokine} levels and \textbf{radiographic progression} in patients with \textbf{ankylosing spondylitis} &leptin; rheumatic diseases; adiponectin; necrosis; DKK-1; IL-6-RFP\\\hline
\end{tabularx}
\caption{\label{table:newidea}More Link Prediction Examples (bold words are entities detected from titles)}
\end{table*}

\begin{table}[!htb]
\small
\begin{tabularx}{\linewidth}{|>{\hsize=0.8\hsize}X|>{\centering\arraybackslash\hsize=1\hsize}X|>{\centering\arraybackslash\hsize=1.6\hsize}X|>{\centering\arraybackslash\hsize=0.6\hsize}X|}\hline
                    & \textbf{Abstract} & \textbf{Conclusion and Future Work} & \textbf{Title} \\\hline
System & 112.4    & 88.1       & 16.5  \\\hline
Human    & 106.5    & 105.5       & 13.0 \\\hline
\end{tabularx}
\caption{\label{table:length} The Average Number of Words of System and Human Output}
\end{table}

\begin{table}[!htb]
\small
\centering
    \begin{tabular}{|c|c|c|c|c|c|}
    \hline
     Output    & \bf 1  & \bf 2  &\bf 3 &\bf 4  &\bf 5  \\ \hline
    \bf Abstracts & 58.3 & 20.1 & 8.03 & 3.60 & 1.46 \\ \hline
    \bf Conclusions & 43.8 & 12.5 & 5.52 & 2.58 & 1.28 \\ \hline
    \bf Titles & 20.1 & 1.31 & 0.23 & 0.06 & 0.00 \\ \hline
\end{tabular}
\caption{\label{table:plag_check}Plagiarism Check: Percentage (\%) of $n$-grams in human input which appear in system generated output for test data.}
\end{table}

\subsection{Analysis and Discussions}

To better justify the function of each component, we conduct ablation studies by removing memory networks, link prediction, and repetition removal respectively. The results are shown in Table~\ref{table:analysis_walk}. We can see that the approach without memory networks tends to diverge from the main topic, especially for generating long texts such as abstracts (the detailed length statistics are shown in Table~\ref{table:length}). From Table~\ref{table:analysis_walk} we can see the later parts of the abstract (Methods and Results) include topically irrelevant entities such as ``\emph{imipramine}"  which is used to treat depression instead of human prostate cancer.

Link prediction successfully introduces new and topically related ideas, such as ``\emph{RT-PCR}" and ``\emph{western blot}" which are two methods for analyzing the expression level of Snail protein, as also mentioned in the human written abstract in Table~\ref{table:walkthrough}. Table~\ref{table:newidea} shows more examples of entities which are related to the entities in input titles based on link prediction. We can see that the predicted entities are often genes or proteins which cause the disease mentioned in a given title, or other diseases from the same family.

Our simple beam search based masking method successfully removes some repeated words and phrases and thus produces more informative output.
The plagiarism check in Table~\ref{table:plag_check} shows our model is creative, because it's not simply copying from the human input.

\subsection{Remaining Challenges}

Our generation model is still largely dependent on language model and extracted facts, and thus it lacks of
knowledge reasoning. It generates a few incorrect abbreviations such as \emph{``Organophosphates(BA)"}, \emph{``chronic kidney disease(UC)"} and \emph{``Fibrosis(DC)"}) because they appear rarely in the training data and thus their contextual representations are not reliable. It also generates some incorrect numbers (e.g., \emph{``The patients were divided into four groups : \textbf{Group 1} , \textbf{Group B}..."}) and pronouns (e.g., ``\emph{A \textbf{63-year-old man} was referred to our hospital ... \textbf{she} was treated with the use of the descending coronary artery"} ).

All of the system generated titles are declarative sentences
while human generated titles are often more engaging (e.g., \emph{``Does HPV play any role in the initiation or prognosis of endometrial adenocarcinomas?"}). Human generated titles often include more concrete and detailed ideas such as \emph{``etumorType , An Algorithm of Discriminating Cancer Types for Circulating Tumor Cells or Cell-free DNAs in Blood"}, and even create new entity abbreviations such as \emph{etumorType} in this example.

\subsection{Requirements to Make PaperRobot Work: Case Study on NLP Domain}

When a cool Natural Language Processing (NLP) system like \emph{PaperRobot} is built, it's natural to ask whether she can benefit the NLP community itself. We re-build the system based on 23,594 NLP papers from the new ACL Anthology Network~\cite{Radev2013}. For knowledge extraction we apply our previous system trained for the NLP domain~\cite{luan2018multi}. But the results are much less satisfactory compared to the biomedical domain.
Due to the small size of data, the language model is not able to effectively copy out-of-vocabulary words and thus the output is often too generic. For example, given a title ``\emph{Statistics based  hybrid approach to Chinese base phrase identification}", \emph{PaperRobot} generates a fluent but uninformative abstract ``\emph{This paper describes a novel approach to the task of Chinese-base-phrase identification. We first utilize the solid foundation for the Chinese parser, and we show that our tool can be easily extended to meet the needs of the sentence structure.}".

Moreover, compared to the biomedical domain, the types of entities and relations in the NLP domain are rather coarse-grained, which often leads to inaccurate prediction of related entities. For example, for an NLP paper title ``\emph{Extracting molecular binding relationships from biomedical text}", \emph{PaperRobot} mistakenly extracts ``\emph{prolog}" as a related entity and generates an abstract ``\emph{In this paper, we present a novel approach to the problem of extracting relationships among the \textbf{prolog}  program. We present a system that uses a macromolecular binding relationships to extract the relationships between the abstracts of the entry. The results show that the system is able to extract the most important concepts in the \textbf{prolog} program.}".

\section{Related Work}

\textbf{Link Prediction.}
Translation-based approaches~\cite{nickel2011three,bordes2013translating,wang2014knowledge,lin2015learning,ji2015knowledge} have been widely exploited for link prediction. Compared with these studies, we are the first to incorporate multi-head graph attention~\cite{sukhbaatar2015end,P18-1136,velickovic2018graph} to encourage the model to capture multi-aspect relevance among nodes. Similar to~\cite{wang2016text,ijcai2017-183}, we enrich entity representation by combining the contextual sentences that include the target entity and its neighbors from the graph structure. This is the first work to incorporate new idea creation via link prediction into automatic paper writing.

\textbf{Knowledge-driven Generation.} Deep Neural Networks have been applied to generate natural language to describe structured knowledge bases~\cite{link13,konstas2013global,flanigan2016generation,D18-1086,kk2,P18-1151,D18-1112,P18-1136,D18-1422},
biographies based on attributes~\cite{biogen16, 17onebio,table2text17,sha2017order,kaffee2018learning,W18-6502,D18-1356}, and image/video captions based on background entities and events~\cite{krishnamoorthy2013generating,wu2018image,D18-1433,D18-1435}.
To handle unknown words, we design an architecture similar to pointer-generator networks~\cite{hybridp17} and copy mechanism~\cite{copy16}. Some interesting applications include generating abstracts based on titles for the natural language processing domain~\cite{P18-2042}, generating a poster~\cite{qiang2016learning} or a science news blog title~\cite{Vadapalli2018} about a published paper.
This is the first work on automatic writing of key paper elements for the biomedical domain, especially conclusion and future work, and follow-on paper titles.

\section{Conclusions and Future Work}
We build a \emph{PaperRobot} who can predict related entities for an input title and write some key elements of a new paper (abstract, conclusion and future work) and predict a new title.  Automatic evaluations and human Turing tests both demonstrate her promising performance.
\emph{PaperRobot} is merely an assistant to help scientists speed up scientific discovery and production. Conducting experiments is beyond her scope, and each of her current components still requires human intervention: constructed knowledge graphs cannot cover all technical details, predicted new links need to be verified, and paper drafts need further editing.
In the future, we plan to develop techniques for extracting entities of more fine-grained entity types, and extend \emph{PaperRobot} to write related work, predict authors, their affiliations and publication venues.

\section*{Acknowledgments}
The knowledge extraction and prediction components were supported by the U.S. NSF No. 1741634 and Tencent AI Lab Rhino-Bird Gift Fund. The views and conclusions contained in this document are those of the authors and should not be interpreted as representing the official policies, either expressed or implied, of the U.S. Government. The U.S. Government is authorized to reproduce and distribute reprints for Government purposes notwithstanding any copyright notation here on.

\appendix
\section{Hyperparameters}
\label{sec:appendix}
Table~\ref{tab:Hyperparameter} shows the hyperparameters of our model.
\begin{table}[htb!]
\setlength\tabcolsep{4pt}
\setlength\extrarowheight{3pt}
\small
\centering
\begin{tabularx}{\linewidth}{|>{\hsize=0.6\hsize}X|>{\hsize=1.1\hsize}X|>{\centering\arraybackslash\hsize=1.3\hsize}X|}
\hline
\textbf{Models}&\textbf{Parameter}                       & \textbf{Value} \\ \hline
\multirow{5}{*}{\parbox{\dimexpr 1\hsize}{Link\\Prediction}}&\# Multi-head                     & 8            \\  \cline{2-3}
&Multi-head hidden                   & 8            \\ \cline{2-3}
&Entity embedding  & 64            \\ \cline{2-3}
&LeakyReLU $\alpha$          & 0.2              \\ \cline{2-3}
&Margin loss $\gamma$ &1\\\hline
\multirow{2}{*}{\parbox{\dimexpr 1\hsize}{Paper Writing}}&Decoder hidden                      & 256            \\ \cline{2-3}
&Coverage loss $\lambda$          & 1              \\ \hline
\multirow{4}{*}{\parbox{\dimexpr 1\hsize}{Both}}&Vocabulary size               & 32553          \\ \cline{2-3}
&Text embedding  & 128            \\ \cline{2-3}
&Optimization                           & Adam \citep{kingma2014adam}           \\ \cline{2-3}
&Learning rate                           & 0.001          \\ \hline
\end{tabularx}
\caption{Hyperparameters\label{tab:Hyperparameter}}
\end{table}

\section{Language model details}
The perplexity scores of the language model are 96.24, 59.69, and 120.31 on titles, abstracts, and conclusions respectively.
\onecolumn
\section{Good Examples that Passed Turing Test}

\begin{table*}[htb!]
\begin{tabularx}{\linewidth}{|>{\hsize=1\hsize}X|>{\hsize=1\hsize}X|}\hline
\textbf{System Output}&\textbf{Human Output}\\\hline
Efficacy and Safety of Artesunate in the Treatment of Uncomplicated Malaria: a Systematic Review and Meta-analysis.
&Low RBM3 Protein Expression Correlates with Clinical Stage, PrognOStic Classification and Increased Risk of Treatment Failure in Testicular Non-Seminomatous Germ Cell Cancer.
\\\hline
Background The aim of the present study was to investigate the effect of Cnidium Lactone on the expression of Mutant and histone deacetylase (HDAC) inhibitors in human prostate cancer (PC). Material/Methods We evaluated the effects of Prostate Cancer on cell proliferation and invasion in vitro and in vivo. Cells were incubated with a single dose of 25 (50 mg/kg), and 10 (100 mg/kg/day), respectively. The primary endpoint was the ability of the mRNA and protein levels of transcription factor (VEGF).
&Background Cnidium Lactone is a natural coumarin compound that can inhibit a variety of cancer cell proliferation and induce cancer cell apoptosis. This experiment investigated the effect of cnidium Lactone on molecular marker expression in prostate cancer nude mice to study its effect in inducing apoptosis . Material/Methods We randomly and equally divided 30 male BALB/C nude mice inoculated with human prostate cancer cells PC-3 into a negative control group, a cyclophosphamide group (500 mg/Kg) , and cnidium Lactone groups at 3 doses (280 mg/Kg, 140 mg/Kg, and 70 mg/Kg). The mice were weighed at 2 weeks after administration.

\\\hline
Abstract Rationale: Aliskiren is a rare disease characterized by a variety of hypertensive disorders . The aim of the present study was to evaluate the effectiveness of aliskiren , pharmacodynamics , and clinical outcomes in patients with hypertension . Methods We reviewed the medical records of ambulatory blood pressure ( BP ) , kinetics , and high-sensitivity C-reactive protein ( CRP ) levels in the treatment of corneal tissue . We performed a retrospective review of the English literature search of PubMed , EMBASE , and Cochrane Library databases . The primary outcome was established by using a scoring system.
&The use of nanoparticles in medicine is an attractive proposition. In the present study, Zinc oxide and silver nanoparticles were evaluated for their antidiabetic activity . Fifty male albino rats with weight 120 $\pm$ 20 and age 6 months were used . Animals were grouped as follows: control; did not receive any type of treatment, diabetic; received a single intraperitoneal dose of streptozotocin (100 mg/kg), diabetic + Zinc oxide nanoparticles (ZnONPs), received single daily oral dose of 10 mg/kg ZnONPs in suspension, diabetic + silver nanoparticles (SNPs); received a single daily oral dose of SNP of 10 mg/kg in suspension and diabetic + insulin; received a single subcutaneous dose of 0.6 units/50 g body.
\\\hline
\end{tabularx}
\caption{Good Examples that Passed Turing Test}
\end{table*}

\begin{table*}[htb!]
\begin{tabularx}{\linewidth}{|>{\hsize=1\hsize}X|>{\hsize=1\hsize}X|}\hline
\textbf{System Output}&\textbf{Human Output}\\\hline
In conclusion , our study demonstrated that HOTAIR transcript expression in NSCLC cells. These results suggest that the overexpression of metastasis may play a role in regulating tumor progression and invasion. Further studies are needed to elucidate the molecular mechanisms involved in the development of cancer.
&VWF is an autocrine/paracrine effector of signal transduction and gene expression in ECs that regulates EC adhesiveness for MSCs via activation of p38 MAPK in ECs.
\\\hline
In summary, the present study demonstrated that BBR could suppress tubulointerstitial fibrosis in NRK 52E cells. In addition, the effects of action on the EMT and HG of DN in the liver cell lines, and the inhibition of renal function may be a potential therapeutic agent for the treatment of diabetic mice. Further studies are needed to elucidate the mechanisms underlying the mechanism of these drugs in the future.
&We characterised KGN cells as a malignant tumour model of GCTs. Continuously cultivated KGN cells acquire an aggressive phenotype, confirmed by the analysis of cellular activities and the expression of biomarkers. More strikingly, KGN cells injected under the skin were metastatic with nodule formation occurring mostly in the bowel. Thus, this cell line is a good model for analysing GCT progression and the mechanisms of metastasis.
\\\hline
In summary, the present study demonstrated that Hydrogen alleviates neuronal apoptosis in SAH rats. These results suggest that the Akt/GSK3β signaling pathway may be a novel therapeutic target for the treatment of EBI.
&In reproductive-age women with ovarian endometriosis, the transcriptional factor SOX2 and NANOG are over expression. Future studies is need to determine their role in pathogenesis of ovarian endometriosis.
\\\hline
In conclusion, the present study demonstrated that DNA methylation and BMP-2 expression was associated with a higher risk of developing Wnt/β-catenin pathway in OA chondrocytes. These results suggest that the SOST of Wnt signaling pathways may be a potential target for the treatment of disease.
&Our novel data strongly suggest that BMP-2 signaling modulates SOST transcription in OA through changes in Smad 1/5/8 binding affinity to the CpG region located upstream of the TSS in the SOST gene, pointing towards the involvement of DNA methylation in SOST expression in OA.\\\hline
The role of cancer stem cells to trastuzumab-based and breast cancer cell proliferation, migration, and invasion.
&Long-term supplementation of decaffeinated green tea extract does not modify body weight or abdominal obesity in a randomized trial of men at high risk for Prostate cancer.\\\hline
\end{tabularx}
\caption{Good Examples that Passed Turing Test}
\end{table*}
\clearpage
\twocolumn

\bibliography{acl2019}
\bibliographystyle{acl_natbib}

\end{document}